%% file: main.tex
\documentclass[lettersize,journal]{IEEEtran}
\usepackage{amsmath,amsfonts}
\usepackage{algorithmic}
\usepackage{array}
\usepackage{textcomp}
\usepackage{stfloats}
\usepackage{url}
\usepackage{verbatim}
\usepackage{graphicx}
\usepackage{multirow}

\usepackage{hyperref} %
\hypersetup{colorlinks=true,linkcolor=blue,anchorcolor=blue,citecolor=blue}
\usepackage{booktabs}

\usepackage{colortbl}
\definecolor{myorange}{RGB}{197,90,17}
\definecolor{gray}{RGB}{200, 200, 200} 

\def\BibTeX{{\rm B\kern-.05em{\sc i\kern-.025em b}\kern-.08em
    T\kern-.1667em\lower.7ex\hbox{E}\kern-.125emX}}
\usepackage{balance}
\begin{document}
\title{D$^2$GSLAM: 4D Dynamic Gaussian Splatting SLAM}
\author{Siting Zhu, Yuxiang Huang, Wenhua Wu, Chaokang Jiang, Yongbo Chen, I-Ming Chen~\IEEEmembership{Fellow,~IEEE}, Hesheng Wang~\IEEEmembership{Senior Member,~IEEE}
\thanks{\quad This work was supported in part by the Natural Science Foundation of China under Grant 62225309, U24A20278, 62361166632, U21A20480 and 62403311. Corresponding Author: Hesheng Wang (email: wanghesheng@sjtu.edu.cn)}
\thanks{\quad Siting Zhu, Wenhua Wu, and Yongbo Chen are with the Department of Automation, Shanghai Jiao Tong University, China. Yuxiang Huang is with the School of Information Science and Electronic Engineering, Shanghai Jiao Tong University, China.
Chaokang Jiang is with Bosch, China.
I-Ming Chen is with the School of Mechanical and Aerospace Engineering, Nanyang Technological University, Singapore.
Hesheng Wang is with Department of Automation, Key Laboratory of System Control and Information Processing of Ministry of Education, State Key Laboratory of Avionics Integration and Aviation System-of-Systems Synthesis, Shanghai Key Laboratory of Navigation and Location Based Services, Shanghai Jiao Tong University, Shanghai, China. (email: \{zhusiting, crosshill, 2142431167wwh, shjtdx\_cyb\}@sjtu.edu.cn, ts20060079a31@cumt.edu.cn, michen@ntu.edu.sg) }
}

% \markboth{Journal of \LaTeX\ Class Files,~Vol.~18, No.~9, September~2020}%

\maketitle

\begin{abstract}
Recent advances in Dense Simultaneous Localization and Mapping (SLAM) have demonstrated remarkable performance in static environments. However, dense SLAM in dynamic environments remains challenging. Most methods directly remove dynamic objects and focus solely on static scene reconstruction, which ignores the motion information contained in these dynamic objects. In this paper, we present D$^2$GSLAM, a novel dynamic SLAM system utilizing Gaussian representation, which simultaneously performs accurate dynamic reconstruction and robust tracking within dynamic environments. Our system is composed of four key components: (i)
We propose a geometric-prompt dynamic separation method to distinguish between static and dynamic elements of the scene. 
This approach leverages the geometric consistency of Gaussian representation and scene geometry to obtain coarse dynamic regions. The regions then serve as prompts to guide the refinement of the coarse mask for achieving accurate motion mask.
(ii) To facilitate accurate and efficient mapping of the dynamic scene, we introduce dynamic-static composite representation that integrates static 3D Gaussians with dynamic 4D Gaussians. This representation allows for modeling the transitions between static and dynamic states of objects in the scene for composite mapping and optimization.
(iii) We employ a progressive pose refinement strategy that leverages both the multi-view consistency of static scene geometry and motion information from dynamic objects to achieve accurate camera tracking.
(iv) We introduce a motion consistency loss, which leverages the temporal continuity in object motions for accurate dynamic modeling. Our D$^2$GSLAM demonstrates superior performance on dynamic scenes in terms of mapping and tracking accuracy, while also showing capability in accurate dynamic modeling. 
\end{abstract}


\input{sec/1_introduction}
\input{sec/2_related}
\input{sec/3_method}

\input{sec/4_experiment}

\input{sec/5_conclusion}

\bibliographystyle{IEEEtran}
\bibliography{IEEEabrv,main}

\end{document}

%% file: sec/1_introduction.tex
\IEEEdisplaynontitleabstractindextext
\IEEEpeerreviewmaketitle

\ifCLASSOPTIONcompsoc
\IEEEraisesectionheading{\section{Introduction}\label{sec:introduction}}
\else
\section{Introduction}
\label{sec:introduction}
\fi

\IEEEPARstart{D}{ense} Simultaneous Localization and Mapping (SLAM) is a fundamental challenge in robotics and artificial intelligence. It aims to jointly perform pose estimation~\cite{yin2024survey} and dense mapping~\cite{zhou2020deeptam, han2022scene} of an unknown environment. Traditional visual SLAM systems~\cite{campos2021orb, kerl2013dense, teed2021droid} have achieved robust tracking and accurate mapping in static environments. However, they fail to perform tracking in dynamic environments due to interference from moving objects. To tackle this issue, existing dynamic SLAM methods~\cite{bescos2018dynaslam, palazzolo2019refusion, scona2018staticfusion, yu2018ds, bescos2021dynaslam, xiao2019dynamic, wang2022drg, hu2022cfp, schorghuber2019slamantic} have attempted to enhance tracking performance by identifying and removing dynamic parts from the input observations. However, these methods lead to incomplete static mapping because the regions occupied by dynamic objects are considered unobserved and left unmapped. Consequently, the reconstructed static map contains holes in the regions consistently occupied by moving objects from different views.

Inspired by the continuous scene modeling capability of Neural Radiance Fields (NeRF)~\cite{mildenhall2021nerf}, NeRF-based dynamic SLAM methods~\cite{li2024learn, jiang2024rodyn, li2024ddn, xu2024nid} have been proposed to achieve complete reconstruction of static scene parts. Nevertheless, they either rely on pretrained networks to segment potential dynamic objects or require learning over long sequences for dynamic separation, which is inefficient for real-world applications.  
Furthermore, existing dynamic SLAM systems typically remove dynamic objects and only reconstruct static parts of scenes, which ignores the motion information contained in these dynamic objects. Meanwhile, capturing and modeling this motion information is crucial for achieving a comprehensive understanding of the scene and enabling downstream tasks such as manipulation and navigation in dynamic environments. Therefore, building a dynamic modeling SLAM system is essential.

However, developing such SLAM system is challenging. First, accurate motion modeling requires sufficient multi-view information for optimization. In SLAM tasks, only limited sequential views are available for dynamic modeling, which limits the system's ability to effectively model the dynamic parts of the scene.
Moreover, modeling the entire environment dynamically leads to high computational costs and information redundancy due to unnecessarily processing the motion information of static parts. 
Therefore, a more efficient strategy is to model dynamic and static scene parts separately. However, this strategy poses significant challenge in generating accurate motion masks without scene-specific pretraining and long sequence learning, as the categories of moving objects and their motions in the scene are complex and unknown.

In this paper, we propose D$^2$GSLAM, a novel dynamic SLAM system that utilizes Gaussian representation~\cite{kerbl20233d} to achieve accurate dynamic reconstruction and robust tracking in dynamic environments. We introduce a geometric-prompt dynamic separation method to generate accurate motion masks. This method leverages the multi-view geometry consistency of scene representation to obtain coarse dynamic regions. Based on these regions, prompts are generated to guide the refinement of the coarse mask in producing fine motion masks.
Moreover, we employ a dynamic-static composite representation that incorporates static 3D Gaussians with dynamic 4D Gaussians for efficient dense mapping in dynamic scenes. To avoid catastrophic forgetting in composite mapping, we introduce retrospective frame optimization that continuously revisits and optimizes previous observations for enhanced reconstruction performance.
Additionally, we introduce a motion consistency loss that leverages the temporal continuity of dynamic motions to optimize dynamic modeling from limited sequential views. 
Furthermore, we employ a progressive pose refinement strategy for accurate camera tracking in dynamic scenes. This strategy implements a hierarchical optimization framework that progresses from static-only initialization to joint static-dynamic refinement, and finally leverages multi-view geometric consistency of the scene for global pose optimization.

Our contributions are summarized as follows:
\begin{itemize}
  \item We present a novel dynamic SLAM system that achieves accurate dynamic reconstruction, motion modeling, and robust tracking in dynamic environments. We employ geometric-prompt dynamic separation strategy to generate accurate motion masks for static-dynamic separate modeling.
  \item We introduce dynamic-static composite representation that integrates static 3D Gaussians with dynamic 4D Gaussians for accurate reconstruction of dynamic scenes. In addition, we employ retrospective frame optimization to augment dynamic reconstruction performance.
  Moreover, we introduce a motion consistency loss
 that utilizes the temporal continuity of motions to guide the optimization of dynamic modeling.
  \item We employ a progressive pose refinement strategy to achieve robust camera tracking in dynamic scenes. This strategy performs hierarchical optimization that advances from static-only initialization to joint static-dynamic refinement, and finally to global optimization leveraging multi-view geometric consistency.
  \item Extensive evaluations are conducted on two challenging dynamic datasets and one common static dataset to demonstrate that our method achieves superior performance compared with existing SLAM methods.
\end{itemize}

%% file: sec/2_related.tex
\section{Related work}
Existing visual SLAM methods can be classified into three main categories based on their scene representation: traditional SLAM, neural implicit SLAM, and 3D Gaussian Splatting SLAM approaches. We review these approaches in both static and dynamic scenarios.

\label{sec:related}
\subsection{Traditional SLAM}
Traditional visual SLAM~\cite{joo2021linear} employs explicit 3D representation for dense mapping, such as point cloud~\cite{kerl2013dense, mur2017orb, teed2021droid}, surfels~\cite{whelan2015elasticfusion, schops2019bad}, mesh~\cite{bloesch2019learning}, Truncated Signed Distance Fields (TSDF)~\cite{niessner2013real, dai2017bundlefusion}.
DROID-SLAM~\cite{teed2021droid} designs a differentiable Dense Bundle Adjustment (DBA) layer for updating camera poses and dense per-pixel depth. ElasticFusion~\cite{whelan2015elasticfusion} introduces a windowed surfel map fusion method coupled with frequent model refinement through non-rigid surface deformations for dense mapping. 
These approaches are effective in static environments for dense mapping and tracking, but lack sufficient appearance and geometry constraints in dynamic scenes.

To achieve robust SLAM in dynamic scenes, existing traditional dynamic SLAM methods~\cite{bescos2018dynaslam, palazzolo2019refusion, scona2018staticfusion, yu2018ds, bescos2021dynaslam, xiao2019dynamic, wang2022drg, hu2022cfp, schorghuber2019slamantic, dai2020rgb, zou2012coslam} distinguish and remove dynamic objects.
These approaches primarily employ two strategies: \textit{(i)} utilizing semantic segmentation~\cite{he2017mask} results to separate potential dynamic objects, \textit{(ii)} leveraging inconsistency in motions to identify the regions occupied by dynamic objects. 
DynaSLAM~\cite{bescos2018dynaslam} combines multi-view constraints and segmentation networks to mask dynamic objects. ReFusion~\cite{palazzolo2019refusion} detects dynamics by analyzing the residuals obtained from map registration. 
StaticFusion~\cite{scona2018staticfusion} simultaneously estimates the camera motion as well as a probabilistic static/dynamic segmentation of the current image pair.
However, these methods fail to achieve complete static mapping since the regions consistently occupied by dynamic objects remain unmapped.

\subsection{Neural implicit SLAM}
Neural radiance fields (NeRF)~\cite{mildenhall2021nerf} is a powerful implicit scene representation for 3D scene modeling. It models scenes as implicit continuous functions represented by neural networks, and has shown promising performance in SLAM tasks~\cite{imap, nice, eslam, yang2022vox, zhu2024sni, zhu2025sni} due to its compact scene modeling capability. iMAP~\cite{imap} introduces a single multi-layer perceptron (MLP) network for dense mapping and localization. NICE-SLAM~\cite{nice} adopts hierarchical feature grid as scene representation, enabling more robust mapping.

The above methods only perform accurate SLAM in static scenes and fail to handle dynamic environments. To achieve neural implicit SLAM in dynamic scenes, \cite{li2024learn} employs a continually-learned classifier to separate dynamic objects for robust tracking. RoDyn-SLAM~\cite{jiang2024rodyn} fuses the optical flow masks with semantic masks to achieve dynamic masks.
However, existing methods~\cite{li2024learn, jiang2024rodyn, li2024ddn, xu2024nid} only achieve the removal of dynamic objects for static reconstruction, rather than modeling the dynamic motions in the scene. In robotics applications, it is crucial to understand how dynamic objects move over time in order to accomplish downstream tasks such as manipulation and navigation. 
Some recent methods~\cite{schischka2024dynamon, duan2023tiv} have achieved dynamic reconstruction by integrating dynamic modeling as an offline post-processing module into SLAM frameworks. 
However, these methods require hours of training, making them impractical for robotics applications that demand real-time performance.
In this paper, we propose a dynamic SLAM system that achieves real-time static reconstruction and dynamic modeling.

\begin{figure*}
    \centering
    \includegraphics[width=\linewidth]{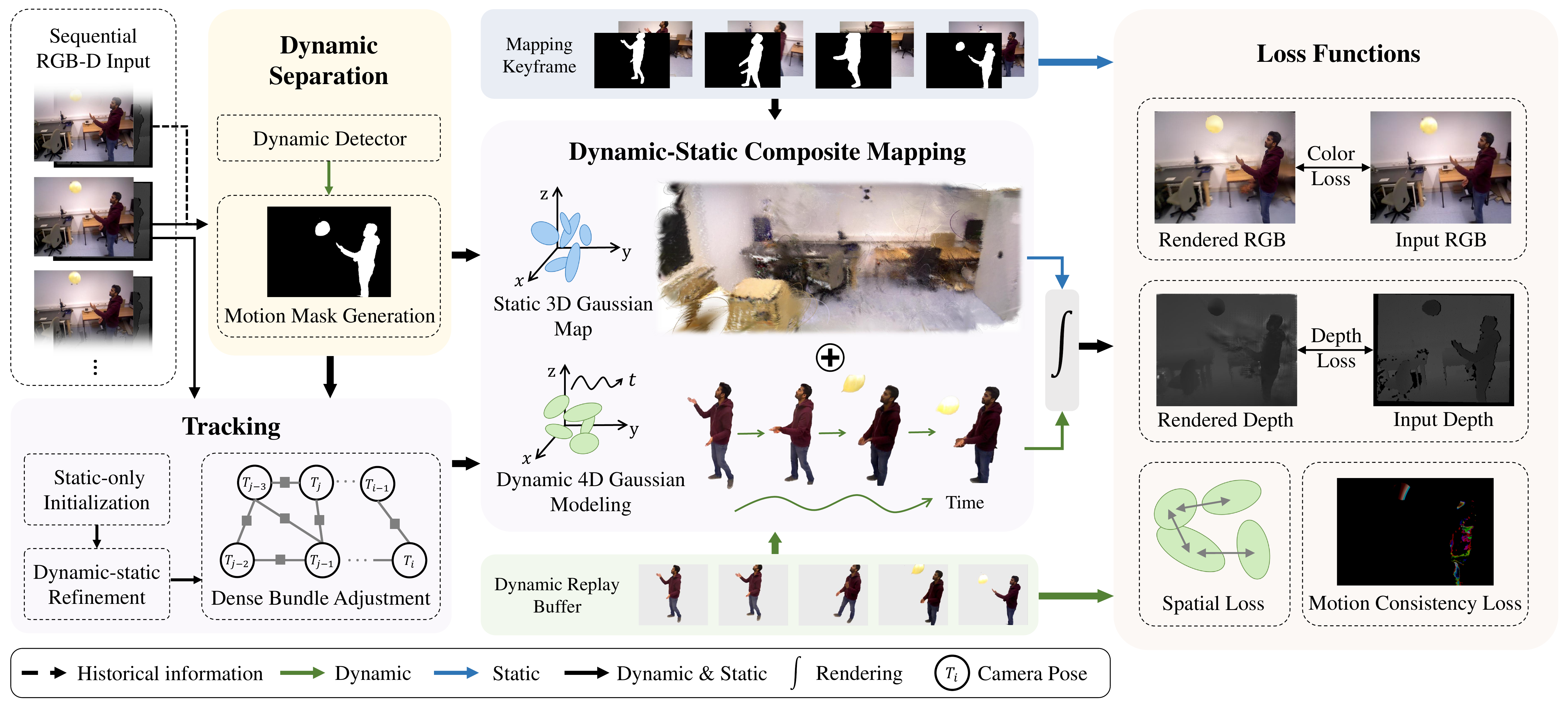}
    \vspace{-0.25in}
    \caption{\textbf{Overview of D$^2$GSLAM.} Our method takes sequential RGB-D frames as input. RGB-D images are first fed into dynamic detector to distinguish whether there are dynamic objects in the current frame. Subsequently, dynamic mask of the current frame is generated through motion mask generation. We perform static-only initialization and dynamic-static refinement, followed by dense bundle adjustment based on generated motion masks to obtain estimated poses in tracking process.
    Then, in dynamic-static composite mapping, we employ dynamic 4D Gaussians and static 3D Gaussians for separate modeling of dynamic and static parts. Concurrently, dynamic replay buffer and mapping keyframe are maintained for retrospective frame optimization, enabling accurate dynamic mapping. }
    \label{fig:overview}
\end{figure*}

\subsection{3D Gaussian Splatting SLAM}
3D Gaussian Splatting (3DGS)~\cite{kerbl20233d, zhu20243d} emerges as a promising radiance field for compact and explicit scene representation. This representation has a wide range of applications, including dynamic reconstruction~\cite{duan20244d, wu20244d, sun20243dgstream, lu20243d, yang2024deformable, lin2024gaussian, yang2023real, li2024spacetime, huang2024sc}, semantic segmentation~\cite{zhou2024feature, ye2023gaussian, wu2024opengaussian}. 
Our main focus is 3DGS-based SLAM~\cite{matsuki2024gaussian, keetha2024splatam, yan2024gs, yugay2023gaussian, ha2025rgbd, peng2024rtg, wang2025mba, zhu2024semgauss}.  
MonoGS~\cite{matsuki2024gaussian} introduces an analytic Jacobian for direct camera pose estimation. 
SplaTAM~\cite{keetha2024splatam} employs silhouette-guided optimization. 
GS-ICP SLAM~\cite{ha2025rgbd} presents a real-time dense representation that combines generalized iterative closest point (ICP)~\cite{segal2009generalized} and 3DGS. Gaussian-SLAM~\cite{yugay2023gaussian} splits the Gaussian map into sub-maps for optimization. 
These methods are only capable of achieving robust tracking and mapping in static scenes. 

Recent works~\cite{xu2025dg, zheng2025wildgs, wen2025gassidy, Hu2017DyGS} remove dynamic objects and reconstruct static parts of dynamic scenes. DG-SLAM~\cite{xu2025dg} leverages pretrained semantic segmentation module to segment dynamic objects, thereby enabling tracking in dynamic scenes. To enable reconstruction of dynamic objects, concurrent works~\cite{matsuki20254dtam, li2025pg} incorporate dynamic Gaussian representation into mapping process. PG-SLAM~\cite{li2025pg} requires first distinguishing between non-rigid and rigid items to perform separate dynamic reconstruction. However, in real-world applications, it is inconvenient to classify these items. 4DTAM~\cite{matsuki20254dtam} employs a warp-field represented by MLP for dynamic modeling. Nevertheless, this approach is limited to short-duration dynamic reconstruction and cannot effectively handle long-term motion sequences, due to the limited memory capacity of MLP network.
Our work proposes a geometric-prompt dynamic separation method to distinguish dynamic objects for robust tracking and introduces dynamic-static composite representation for accurate reconstruction in dynamic scenes.

%% file: sec/3_method.tex
\section{Method}
The overview of our method is shown in Fig.~\ref{fig:overview}. Given a sequence of RGB-D frames $F = \{C_i, D_i\}_{i \in M}$, we perform reconstruction of both dynamic objects and static background, while conducting robust tracking within dynamic environments. Sec.~\ref{sec:Formulation} presents formulation of our method. 
Sec.~\ref{sec:Dynamic Separation} describes geometric-prompt dynamic separation strategy, including dynamic detector and motion mask generation. 
Sec.~\ref{sec:Composite Mapping} introduces dynamic-static composite mapping, including composite map expansion and retrospective frame optimization. 
Sec.~\ref{sec:Camera Tracking} describes the camera tracking process.
Sec.~\ref{sec:Loss Functions} presents the loss functions. 

\subsection{Formulation}
\label{sec:Formulation}
\noindent\textbf{Static 3D Gaussian.}\hspace{5pt} 
We employ a set of Gaussian primitives $\mathcal{G}_{s}$ for static scene modeling. Each $i$-th 3D Gaussian $\mathcal{G}_{s}^{i}$ includes center position $\mu_{s}^{i}=\{x_{s}^{i}, y_{s}^{i}, z_{s}^{i}\}$, covariance $\Sigma_{s}^{i}\in \mathbb{R}^{3\times 3}$, opacity $o_{s}^{i}$, color $c_{s}^{i}$, and is represented by:
\begin{equation}
\mathcal{G}_{s}(\boldsymbol{x}) = e^{-\frac{1}{2}(x-\mu_{s})^T {\Sigma_{s}}^{-1} (x-\mu_{s})},
\end{equation}
where  $\Sigma_{s}$ is parameterized as $\Sigma_{s}=R_{s} S_{s} {S_{s}}^T {R_{s}}^T$ for optimization. $S_{s}=\text{diag}[(s_x){_{s}}, (s_y){_{s}}, (s_z){_{s}}]$ is a scaling matrix and $R_{s}$ represents 3D rotation.

\noindent\textbf{Dynamic 4D Gaussian.}\hspace{5pt} 
We utilize 4D Gaussian for dynamic modeling by adding time dimension $t$ to 3D Gaussian, following \cite{duan20244d, yang2023real}. 4D Gaussian $\mathcal{G}_{d}^i$ is represented by time-varying center position $\mu_{d}^i=\{x_{d}^i, y_{d}^i, z_{d}^i, t_{d}^i\}$, covariance $\Sigma_{d}^i\in \mathbb{R}^{4\times 4}$, opacity $o_{d}^i$, and constant color $c_{d}^i$. The spatial representation of each 4D Gaussian is formulated as:
\begin{equation}
\mathcal{G}_{d}(\boldsymbol{x}) = e^{-\frac{1}{2}(x-\mu_{d})^T {\Sigma_{d}}^{-1} (x-\mu_{d})}.
\end{equation}
Similar to $\Sigma_{s}$ in static 3D Gaussian, $\Sigma_{d}=R_{d} S_{d} {S_{d}}^T {R_{d}}^T$. Time-varying scaling matrix is formulated as $S_{d}=\text{diag}[(s_x){_{d}}, (s_y){_{d}}, (s_z){_{d}}, (s_t){_{d}}]$. 

Different from 3D rotation $R_{s}$, 4D rotation $R_{d}$ is characterized by 4D rotors for high-dimensional rotation modeling, motivated by~\cite{bosch2020n}. Definition of rotors and detailed derivation are shown in the supplementary material.
$\Sigma_{d}$ is parameterized as $\Sigma_{d} = \footnotesize{\begin{pmatrix} A_{3\times 3} & B_{3\times 1} \\ B^T & C_{1\times 1} \end{pmatrix}}$. Given time $t$, 4D Gaussian can be projected into 3D Gaussian representation $\mathcal{G}_s(\boldsymbol{x},t)$ of this time:
\begin{equation}
\mathcal{G}_s(\boldsymbol{x},t)=e^{-\frac{1}{2}\lambda(t-t_{d})^2}e^{-\frac{1}{2}([x-\mu(t)]^T{\Sigma_{s}}^{-1}[x-\mu(t)])},
\end{equation}
where
\begin{equation}
\begin{gathered}
\lambda = C^{-1}, \
\Sigma_{s} = A - \frac{BB^T}{C}, \\
\mu(t) = (x_{d}, y_{d}, z_{d})^T + (t -  t_{d})\frac{B}{C}.
 \end{gathered}
\end{equation}
$\frac{B}{C}$ represents the motion velocity of time $t$. 

\noindent\textbf{Hybrid Rendering.}\hspace{5pt} 
Our dynamic SLAM system employs $\mathcal{G}=\{\mathcal{G}_{s}^{i_s}, \mathcal{G}_{d}^{i_d}\}_{i_s \in N_s, i_d \in N_d}$ for static reconstruction and dynamic modeling, while utilize hybrid rendering strategy to obtain rendered images. Given time $t$, 4D Gaussian is first projected to 3D Gaussian. Then, rendered pixel color $C_p$ from static 3D Gaussian and projected 4D Gaussian can be calculated by blending Gaussians sorted in depth order:
\begin{equation}
\begin{aligned}
C_p(t) =  \sum_{i_d \in N_d} c_d^{i_d} \alpha_{d}^{i_d}(t) \prod_{j=1}^{i_d-1} (1 - \alpha_{d}^j(t)) + \\
\sum_{i_s \in N_s} c_s^{i_s} \alpha_s^{i_s} \prod_{j=1}^{i_s-1} (1 - \alpha_{s}^j),
\end{aligned}
\end{equation}
where $\alpha^{i}=o^i\mathcal{G'}^i$, $\mathcal{G'}^i$ is 2D projection of $\mathcal{G}^i$. Similarly, depth $D_p(t)$ is rendered through:
\begin{equation}
\begin{aligned}
D_p(t) = \sum_{i_d \in N_d} d_d^{i_d} \alpha_{d}^{i_d}(t) \prod_{j=1}^{i_d-1} (1 - \alpha_{d}^j(t)) + \\
\sum_{i_s \in N_s} d_s^{i_s} \alpha_s^{i_s} \prod_{j=1}^{i_s-1} (1 - \alpha_{s}^j).
\end{aligned}
\end{equation}

\begin{figure*}
    \centering
    \includegraphics[width=\linewidth]{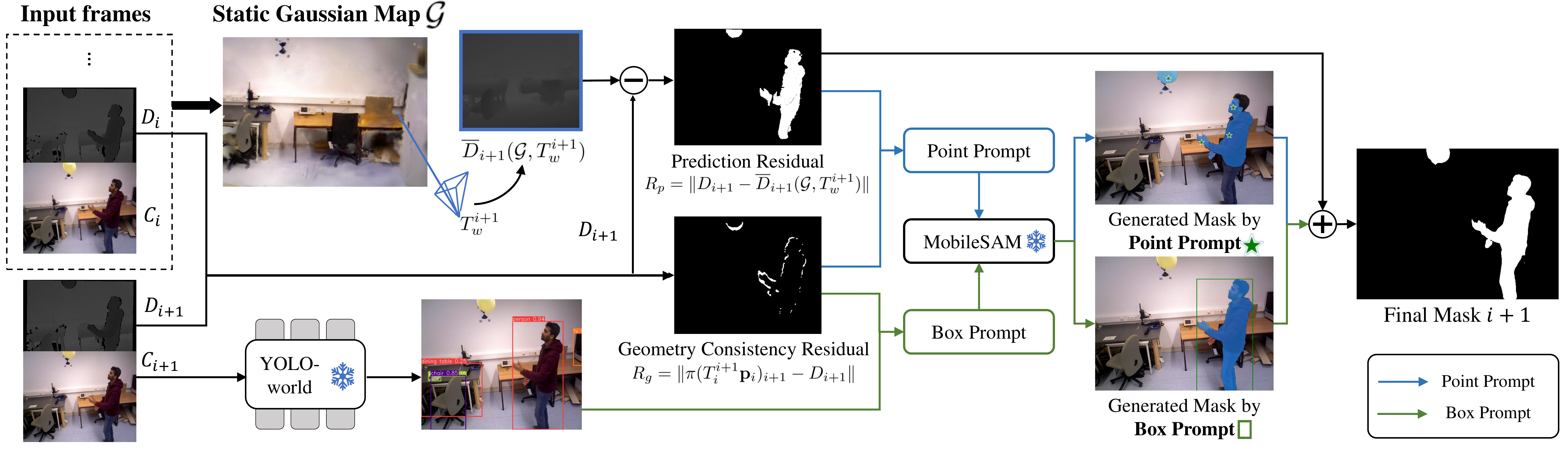}
    \caption{\textbf{Geometric-Prompt dynamic separation.} 
    The box prompt obtained through YOLO-world~\cite{cheng2024yolo} fails to detect the balloon, as 'balloon' is not included in its predefined semantic class categories. By applying our proposed $R_g$ and $R_p$, we successfully obtain comprehensive motion masks for all dynamic objects in the scene. Our method can segment dynamic objects by leveraging scene geometry information, regardless of object categories.
    }
    \label{fig:method_mask}
\end{figure*}
\subsection{Geometric-Prompt Dynamic Separation}
\label{sec:Dynamic Separation}
During the SLAM process, we first employ an efficient dynamic detector to determine whether the input frame contains dynamic objects. Then, we separate dynamic objects from the scene through motion mask generation. Fig.~\ref{fig:method_mask} shows an overview of our geometric-prompt dynamic separation module.

\subsubsection{Dynamic Detector}
\label{sec:Dynamic Detector}
In static environments, the geometric representation of a scene remains invariant across multiple viewpoints. When dynamic objects appear, they introduce spatial inconsistencies in this representation, as they occupy different positions in 3D space between consecutive frames.
We exploit such inconsistency to detect dynamic objects. Specifically, we introduce a geometry consistency residual: 
\begin{equation}  
\label{eq7}
R_g = \|\pi (T_{i}^{i+1}\mathbf{p}_{i})_{i+1} - D_{i+1}\|,
\end{equation}
 $\mathbf{p}_{i}$ is the 3D point cloud obtained by back-projecting the depth map $D_{i}$ of last input frame,
$T_{i}^{i+1}$ denotes relative pose transformation, $\pi ()_{i+1}$ signifies the projection of transformed point cloud onto frame $F_{i+1}$. Due to noise in the input depth and errors in the estimated pose $T_{i}^{i+1}$, identifying dynamic regions directly from $R_g$ may lead to false detections. 

To improve the robustness of dynamic detection, we apply connected component extraction to $R_g$, denoted as $\mathbb{C}_g[]$, and obtain the connected components $\mathbb{C}_g[R_g]$.
Then, we distinguish dynamic objects by comparing the area of connected components: $\mathbb{C}_g[R_g] > \sigma_d$, where $\sigma_d$ controls the sensitivity of dynamic detection. $\mathbb{C}_g[R_g]$ with area larger than $\sigma_d$ are retained as $\{\mathbb{C}_g^{k}\}_{k \in N_{i+1}}$. If $\{\mathbb{C}_g^{k}\}$ is not an empty set, it indicates that there are inconsistencies in the scene. Since the geometry representation of a static scene is consistent across multiple views, we consider that dynamic objects exist in frame $F_{i+1}$. 

\subsubsection{Motion Mask Generation}
\label{sec:Motion Mask Generation}
We first leverage the multi-view geometric consistency property of 3D Gaussian representation to generate coarse dynamic regions. 
Since 3D Gaussians represent scene geometry in a view-independent manner, they enable accurate depth prediction from novel viewpoints for static elements. Considering this property, we predict the depth map $\overline{D}_{i+1}$ of the next frame by rendering the next-view depth information from static Gaussian representation $\mathcal{G}_s$. In this scenario, static regions in this predicted depth map $\overline{D}_{i+1}$ and the observed depth map $D_{i+1}$ exhibit minimal differences, as static Gaussians accurately represent these areas across different views. Conversely, dynamic regions show significant discrepancies. By quantifying these differences, we can effectively locate dynamic objects in the scene. Therefore, we introduce a prediction residual $R_p$ to obtain the potential dynamic regions:
\begin{equation}
\label{eq8}
R_p = \|D_{i+1} - \overline{D}_{i+1}(\mathcal{G}_s, T_w^{i+1})\|,
\end{equation}
where $ \overline{D}_{i+1}(\mathcal{G}_s, T_w^{i+1})$ represents predicted depth map from next-view pose $T_w^{i+1}$. We then binarize $R_p$ using a threshold $\tau_d$ to generate the coarse dynamic regions $R'_p$.

However, these coarse dynamic regions often exhibit discontinuities and imperfections due to errors in Gaussian representation and pose estimation. To obtain more accurate motion masks, we employ a prompt-guided refinement strategy.
Specifically, we generate point prompts by extracting the centroids of connected components in the coarse dynamic regions $R'_p$. Some generated prompts may be incorrectly positioned due to representation errors. Therefore, we filter out prompts that do not intersect with connected components of the geometry consistency residual $\mathbb{C}_g[R_g]$, achieving final point prompts $\mathbb{P}_p$. $\mathbb{P}_p$ will be fed into MobileSAM~\cite{zhang2023faster} to produce object masks indicated by these prompts. Nevertheless, point prompts provide limited spatial guidance, potentially leading to incomplete masks. To enhance mask completeness, we incorporate bounding box prompts. We utilize YOLO-world~\cite{cheng2024yolo} to obtain detection bounding boxes and remove static ones using the same intersection criterion applied to point prompts. These filtered boxes serve as box prompts $\mathbb{P}_b$ of dynamic objects to obtain corresponding object masks. As detection may fail for certain objects, we combine $\mathbb{P}_p$ with $\mathbb{P}_b$ and input them into MobileSAM~\cite{zhang2023faster} to produce refined motion masks. Final mask is the integration of coarse dynamic regions and refined motion masks, as shown in Fig.~\ref{fig:method_mask}. 

Note that in our SLAM system, for each newly input frame $F_{i+1}$, we first detect dynamic regions before performing initial tracking in the static areas to avoid interference from moving objects. As a result, $T_{i}^{i+1}$ (Eq.~\ref{eq7}) and $T_w^{i+1}$ (Eq.~\ref{eq8}) cannot be obtained directly through tracking on frame $F_{i+1}$. To estimate these poses, we assume that the camera does not undergo extreme motion continuously within the scene. Given tracking poses of previous frames $T_w^{i-1}$ and $T_w^{i}$, we predict the relative pose $T_{i}^{i+1} = T_w^{i} (T_w^{i-1})^{-1} $ and the camera pose $T_w^{i+1} = T_{i}^{i+1} T_w^{i}$ for dynamic detection and motion mask generation.

Overall, our geometric-prompt dynamic separation module utilizes scene geometric information to identify dynamic objects. This method does not rely on predefined object categories and requires no prior knowledge of whether objects are rigid or non-rigid. Instead, it takes effect by detecting geometric inconsistencies across sequential frames. When such inconsistencies are present in the scene, indicating object motion or deformation, we can derive corresponding motion masks.

\subsection{Dynamic-Static Composite Mapping}
\label{sec:Composite Mapping}
Our dynamic SLAM system utilizes dynamic-static composite representation that integrates static 3D Gaussians and dynamic 4D Gaussians for dense mapping. During the mapping process, we perform composite map expansion based on RGB-D input and motion masks for incremental mapping. Furthermore, we employ retrospective frame optimization to avoid catastrophic forgetting of previous observations, as well as enhance dynamic modeling capability. 

\subsubsection{Composite Map Expansion}
Given an RGB-D input, we first initialize the SLAM system using the first frame $F_0$. Since determining dynamic objects requires multi-frame temporal information, all objects in $F_0$  are initially reconstructed using static 3D Gaussians. Subsequently, for each input frame, we employ dynamic detector (Sec.~\ref{sec:Dynamic Detector}) to determine whether the frame contains dynamic objects. If no dynamic objects are detected, the SLAM system continues to expand the map of the static scene by adding new 3D Gaussians. 
When dynamic objects are detected in $i$-th frame $F_i$, we generate a motion mask of this frame (Sec.~\ref{sec:Motion Mask Generation}). 
The dynamic objects in $F_i$ fall into two categories: \textit{(i)} newly introduced objects that have not appeared in the scene before, or \textit{(ii)} previously static objects that have become moving, whose Gaussian representation already exists in the global Gaussian map. 

Then, we employ a k-nearest neighbor (KNN) search to determine the status of dynamic objects and apply different map expansion strategies accordingly. To achieve KNN search for map expansion, we first extract 3D point cloud representation of dynamic objects in $F_i$. Based on the motion mask of $F_i$ and input depth $D_i$, we obtain the 3D positions of dynamic objects in pixel coordinates, which constitute a 3D point cloud. Subsequently, using camera intrinsic and camera pose, we transform this 3D point cloud into world coordinates, denoted as $\mathbf{p}_w^i$. We then perform KNN search using $\mathbf{p}_w^i$ as query points within the global Gaussian map to identify corresponding Gaussians $g_c$. If $g_c$ is found, we extend $g_c$, which was previously static 3D Gaussians, by adding a time dimension to obtain dynamic 4D Gaussian representation. When no corresponding Gaussian is found, the dynamic objects are considered new components of the scene. Therefore, 4D Gaussian representation of these objects is directly added to the global map. 
As these dynamic objects progressively enter the field of view, we continuously incorporate new dynamic 4D Gaussians into the global map, enabling incremental 4D Gaussian mapping. Additionally, Gaussians with low opacity will be pruned during mapping.

\subsubsection{Retrospective Frame Optimization}
In the SLAM process, the optimization of 3D Gaussian representation is typically performed based on incoming frames. However, this current frame-based optimization strategy causes the 4D Gaussian representation to overfit the motion information of the latest observation, leading to catastrophic forgetting of previous motions. Meanwhile, dynamic object modeling requires temporal constraints and optimization to accurately model continuous motion patterns.
To address this issue, we introduce and maintain a dynamic replay buffer that contains dynamic motion information from previous time steps. During the mapping process, we continuously backtrack and incorporate previous motion information in this buffer to iteratively optimize dynamic 4D Gaussians. This strategy enables temporally coherent and accurate dynamic modeling within our dynamic SLAM system.

For the static parts of the scene, we select the frames with sufficient relative motion compared to the previous keyframe as mapping keyframes, which are used for map expansion. To enhance the accuracy of scene reconstruction, we employ keyframe retrospective optimization on the static parts. 
Note that we perform more iterations of retrospective optimization in dynamic modeling than in static reconstruction, as dynamic modeling involves higher temporal complexity.
By integrating retrospective optimization for both dynamic and static components, our approach significantly enhances the dynamic modeling and static reconstruction capabilities of our dynamic SLAM system.

\subsection{Camera Tracking}
\label{sec:Camera Tracking}
Our SLAM system leverages both static and dynamic scene information to perform camera tracking in dynamic environments. Specifically, the camera pose estimation is achieved through a progressive refinement process. We first estimate the camera pose using only static scene components, as dynamic objects may not be accurately represented in the initial stages and could adversely affect pose estimation. The initial pose is obtained by minimizing the difference between observed and rendered 2D RGB and depth information from static scene geometry:
\begin{equation}
 \min_{T_w^{i}} \ \mathcal{M}_i(\lambda_c^t \|C(\mathcal{G}_s,T_w^{i}) - C_i\| + \lambda_d^t \|D(\mathcal{G}_s,T_w^{i}) - D_i\|),
\end{equation}
where $\lambda_c^t$ and $\lambda_d^t$ are weighting coefficients, $\mathcal{M}_i$ is motion mask, $C(\mathcal{G}_s,T_w^{i})$ and $D(\mathcal{G}_s,T_w^{i})$ represent the rendered RGB and depth images. 

Subsequently, we refine the estimated pose by incorporating dynamic scene information once it has been sufficiently optimized.
This step integrates additional geometric constraints from dynamic objects, which is essential in dynamic environments where static scene elements alone provide limited observability for reliable pose estimation. Specifically, we render both static and dynamic scene components to construct losses against the observed RGB-D images for optimization. Since the scene representation is initially constructed based on the pose estimate derived from static components, this refinement stage jointly optimizes both camera pose and scene representation to ensure consistency and improve overall accuracy.
The refinement process is formulated as:
\begin{equation}
 \min_{T_w^{i}, G} \ \lambda_c^t \|C(\mathcal{G},T_w^{i},t) - C_i\| + \lambda_d^t \|D(\mathcal{G},T_w^{i},t) - D_i\|,
\end{equation}
where $\mathcal{G}=\{\mathcal{G}_{s}, \mathcal{G}_{d}\}$, $C(\mathcal{G},T_w^{i},t)$ and $D(\mathcal{G},T_w^{i},t)$ represent the rendered RGB and depth images of both static and dynamic components.

Furthermore, to reduce accumulated drift in the tracking process, we perform global bundle adjustment (BA) for pose optimization of keyframes.
This optimization relies only on static scene elements, as dynamic objects inherently lack multi-view consistency across frames.
We maintain a factor graph $(\mathcal{V}, \mathcal{E})$ to establish the co-visibility among frames, where an edge $(i,j)\in\mathcal{E}$ means frame $F_i$ and $F_j$ contain overlapping fields of view. 
Based on this graph, we construct a dense BA optimization built upon \cite{teed2021droid} to optimize the pose $T$:
\begin{equation}
\begin{gathered}
   \arg \min_{T} \sum_{(i,j)\in\mathcal{E}} \|\tilde{p}_{ij} - KT_j^{-1}(T_i(1/d_i)K^{-1}[p_i,1]^T)\|_{\Sigma}^2 \\
   \Sigma = \Sigma_{ij} \oplus \mathcal{M}, \quad \Sigma_{ij} = \mathrm{diag}\,w_{ij},
\end{gathered}
\end{equation}
where $p_i$ represents pixel coordinates, $\tilde{p}_{ij}$ represents the predicted pixel coordinates obtained by projecting the pixel grid from keyframe $i$ into keyframe $j$ using optical flow. $d_i$ is depth of keyframe $i$, $K$ is camera intrinsics. $T_i$ and $T_j$ are camera-to-world extrinsics for keyframes $i$ and $j$ respectively. $w_{ij}$ represents confidence weights. $\mathcal{M}$ is generated motion mask through our geometric-prompt dynamic separation module. We employ $\mathcal{M}$ to the weighted covariance matrix, which suppresses the dynamic parts in optimization process.

\subsection{Loss Functions}
\label{sec:Loss Functions}
We employ various loss functions to optimize our dynamic-static composite representation. 

\noindent\textbf{Color and Depth Loss.}\hspace{5pt}
We construct color loss $\mathcal{L}_{c}$ and depth loss $\mathcal{L}_{d}$ by comparing rendered RGB and depth with input RGB-D values. SSIM term~\cite{kerbl20233d} is added to color loss:
\begin{equation}
\begin{gathered}
 \mathcal{L}_{c} = (1-\lambda)\|C(\mathcal{G},T_w^{i}, t) - C_i\| + \lambda \mathcal{L}_\text{SSIM}, \\
 \mathcal{L}_{d} = \|D(\mathcal{G},T_w^{i},t) - D_i\|,
\end{gathered}
\end{equation}
where $\lambda=0.2$. Given time $t$ and pose $T_w^{i}$, $C(\mathcal{G},T_w^{i}, t)$ and $D(\mathcal{G},T_w^{i},t)$ are rendered RGB and depth from both dynamic and static Gaussians $\mathcal{G}=\{\mathcal{G}_s, \mathcal{G}_d\}$.

\noindent\textbf{Spatial Loss.}\hspace{5pt}
Spatially adjacent dynamic Gaussians should have similar motions.
Therefore, given $i$-th dynamic Gaussian $\mathcal{G}_d^i$, we select $M$ nearest dynamic Gaussians within its neighborhood space $Q_i$ and enforce their motions to be continuous for dynamic optimization:
\begin{equation}
 \mathcal{L}_{sp} = \frac{1}{N_d} \sum_{i=1}^{N_d} \| v_i - \frac{1}{M} \sum_{j \in Q_i} v_j \|,
\end{equation}
where $v_i$ is motion velocity of $\mathcal{G}_d^i$ in time $t$. 
Description of motion velocity is shown in Sec.~\ref{sec:Formulation}.

\noindent\textbf{Motion Consistency Loss.}\hspace{5pt}
In dynamic SLAM, image frames are sequentially fed into the system, resulting in temporal continuity of motions within these frames. We leverage this temporal continuity to construct loss for optimizing the dynamic scene representation.  Specifically, since the motions of the dynamic objects are continuous, we can predict the motion of the next frame $F_{i+1}$ based on historical dynamic modeling information. Then, motion consistency loss is constructed between the prediction and actual observations to optimize Gaussian representation, thereby enhancing the temporal modeling capability of dynamic SLAM. Since the camera frame rate is fixed, we can obtain the motion prediction by projecting 4D Gaussians to 3D Gaussians at time $t_{i+1}$:
\begin{equation}
 \mathcal{L}_{mc} = \|C(\mathcal{G}_d(t_{i+1}),T_{i+1}, t_{i+1}) - C_{i+1}\|,
\end{equation}
where $\mathcal{G}_d(t_{i+1})$ is projected dynamic Gaussians at time $t_{i+1}$, $T_{i+1}$ is estimated camera pose of $F_{i+1}$.

Overall, the complete loss functions for static scene optimization $\mathcal{L}_{static}$ and for dynamic reconstruction $\mathcal{L}_{dynamic}$ are the weighted sum of the above losses:
\begin{equation}
\begin{gathered}
 \mathcal{L}_{static} = \lambda_{c}\mathcal{L}_{c} + \lambda_{d}\mathcal{L}_{d}, \\
  \mathcal{L}_{dynamic} = \lambda_{c}\mathcal{L}_{c} + \lambda_{sp}\mathcal{L}_{sp} + \lambda_{mc}\mathcal{L}_{mc}.
 \end{gathered}
\end{equation}

%% file: sec/4_experiment.tex
\begin{table*}[t]
  \centering
 \caption{Comparison of our method with SLAM baselines in tracking performance.  We report ATE RMSE [cm]~$\downarrow$ on \textbf{11 high-dynamic scenes} of Bonn dataset. (DM) methods that perform dynamic modeling. 
 }
  \resizebox{1.0\linewidth}{!}{
  \begin{tabular}{l|l |c |c|c |c|c |c|c |c|c |c|c |c}
    \toprule
    & Methods & DM & ball & ball2 & ball\_tk & ball\_tk2 & crowd & crowd2 & crowd3 & ps\_tk &  ps\_tk2 & syn &  r\_box\\
    \midrule
    \multirow{4}{*}{\shortstack{Static\\SLAM}}
    & SplaTAM*~\cite{keetha2024splatam} & & 35.0 & 36.8 & 12.5 & 12.6 & 136.6 & 52.7 & 42.2 & 133.4 & 91.2 & 61.2 & 15.4\\
    & MonoGS*~\cite{matsuki2024gaussian}& & 37.1 & 62.5 & 4.3 & 22.8 & 24.2 & 136.3 & 94.5 & 38.8 & 114.3 & 62.4 & 52.9\\
    & GS-ICP SLAM*~\cite{ha2025rgbd} & & 18.5 & 14.6 & 27.9 & 57.14 & 122.5 & 160.5 & 113.8 & 69.4 & 62.7 & 54.7 & 29.5 \\
    & DROID-SLAM*~\cite{teed2021droid} & & 10.4 & 9.7 & 3.5 & 2.9 & 6.7 & 7.2 & 6.1 & 3.6 & 6.1 & 0.6 & 18.9\\
     \midrule
    \multirow{9}{*}{\shortstack{Dynamic\\SLAM}} 
    & ReFusion~\cite{palazzolo2019refusion} & &17.5 & 25.4 & 30.2 & 32.2 & 20.4 & 15.5 & 13.7 & 28.9 & 46.3 & 44.1 & 22.2 \\
    & StaticFusion~\cite{scona2018staticfusion} & & 23.3 & 29.3 & 22.1 & 36.6 & 358.6 & 21.5 & 16.8 & 48.4 & 62.6 & 44.6 & 33.4 \\
    & DynaSLAM~\cite{bescos2018dynaslam}  & & 3.0 & 2.9 & 4.9 & 3.5 & 1.6 & 3.1 & 3.8 & 6.1 & 7.8 & 1.5 & 29.1 \\
    & DDN-SLAM~\cite{li2024ddn}& & \cellcolor{gray}\textbf{1.8} & 4.1 & -- & -- & 1.8 & \cellcolor{gray}\textbf{2.3} & -- & 4.3 & 3.8 & -- & --\\
    & RoDyn-SLAM*~\cite{jiang2024rodyn} & & 7.9 & 11.5 & 13.3 & 30.1 & 12.4 & 12.8 & 13.6 & 14.5 & 13.8 & 50.5 & 25.5 \\
    & LMF-SLAM~\cite{li2024learn} & & 20.6 & 13.6 & -- & -- & 11.6 & 20.0 & 10.7 & 27.4 & -- & 13.0 & --\\
    & DG-SLAM~\cite{xu2025dg} & & 3.7 & 4.1 & 10.0 & -- & -- & -- & -- & 4.5 & 6.9 & -- & -- \\
    & PG-SLAM~\cite{li2025pg} & \checkmark & 6.4 & 7.3 & -- & -- & -- & -- & -- & 5.0 & 8.5 & -- & -- \\
    & DynaMoN~\cite{schischka2024dynamon} & \checkmark & 2.8 & 2.7 & 3.4 & 3.2 & 3.5 & 2.8 & 3.2 & 14.8 & 2.2 & 0.7 & 17.7\\
    & D$^2$GSLAM (Ours) & \checkmark & \cellcolor{gray}\textbf{1.8} & \cellcolor{gray}\textbf{2.2} & \cellcolor{gray}\textbf{3.0} & \cellcolor{gray}\textbf{1.9} & \cellcolor{gray}\textbf{1.4} & \cellcolor{gray}\textbf{2.3} & \cellcolor{gray}\textbf{3.0} & \cellcolor{gray}\textbf{2.5} & \cellcolor{gray}\textbf{2.1} & \cellcolor{gray}\textbf{0.5} & \cellcolor{gray}\textbf{13.8}\\
    \toprule
     \end{tabular}}
  \label{tab:bonn_high}
    \vspace{0.5em} 
    \tiny
    \raggedright 
    \textbf{*} denotes the reproduced results by running official code.  Results of other methods are directly taken from their papers. \textbf{--} denotes that no results were reported for these sequences. Best results are highlighted as \colorbox{gray}{\textbf{first}}. 
\end{table*}

\begin{table*}[t]
  \centering
    \caption{Comparison of our method with SLAM baselines in tracking performance.  We report ATE RMSE [cm]~$\downarrow$ on \textbf{11 low-dynamic scenes} of Bonn dataset. (DM) methods that perform dynamic modeling.  
    }
  \resizebox{1.0\linewidth}{!}{
  \begin{tabular}{l|l |c |c|c |c|c |c|c |c|c |c|c |c}
    \toprule
    & Methods & DM & k\_box & k\_box2 & mv\_n\_box & mv\_n\_box2 & p\_n\_box & p\_n\_box2 & p\_n\_box3 & p\_box &  r\_n\_box & r\_n\_box2 &  syn2\\
    \midrule
    \multirow{4}{*}{\shortstack{Static\\SLAM}}
    & SplaTAM*~\cite{keetha2024splatam} & & 3.5 & 3.83 & 13.9 & 19.0 & 17.6 & 10.4 & 12.6 & 12.2 & 11.8 & 10.7 & 22.7\\
    & MonoGS*~\cite{matsuki2024gaussian}& & 3.9 & 5.2 & 8.9 & 22.9 & 39.1 & 9.2 & 3.1 & 25.2 & 3.5 & 4.3 & \cellcolor{gray}\textbf{0.5}\\
    & GS-ICP SLAM*~\cite{ha2025rgbd} & & 14.3 & 19.0 & 9.1 & 24.8 & 32.4 & 38.9 & 38.3 & 45.0 & 20.3 & 37.7 & 19.4\\
    & DROID-SLAM*~\cite{teed2021droid} & & 2.0 & 1.9 & 2.2 & 4.0 & 7.8 & 2.5 & 2.6 & 12.7 & 1.6 & 2.1 & 1.7\\
     \midrule
     \multirow{9}{*}{\shortstack{Dynamic\\SLAM}}
    & ReFusion~\cite{palazzolo2019refusion} & &14.8 & 16.1 & 7.1 & 17.9 & 10.6 & 14.1 & 17.4 & 57.1 & 4.1 & 11.1 & 2.2\\
    & StaticFusion~\cite{scona2018staticfusion} & & 33.6 & 26.3 & 14.1 & 36.4 & 12.5 & 17.7 & 25.6 & 33.0 & 13.6 & 12.9 & 2.7\\
    & DynaSLAM~\cite{bescos2018dynaslam}  & & 2.9 & 3.5 & 23.2 & 3.9 & 57.5 & 2.1 & 5.8 & 25.5 & 1.6 & 2.1 & 0.9\\
    & DDN-SLAM~\cite{li2024ddn}& & -- & -- & 2.0 & 3.2 & -- & -- & -- & -- & -- & -- & --\\
    & RoDyn-SLAM*~\cite{jiang2024rodyn} & & 11.1 & 14.8 & 7.2 & 12.3 & 9.3 & 9.1 & 10.5 & 16.7 & 17.6 & 9.7 & 35.5\\
    & LMF-SLAM~\cite{li2024learn} & & 11.2 & 10.4 & -- & -- & -- & -- & -- & -- & -- & -- & --\\
    & DG-SLAM~\cite{xu2025dg} & & -- & -- & -- & 3.5 & -- & -- & -- & -- & -- & -- & --\\
    & PG-SLAM~\cite{li2025pg} & \checkmark & -- & -- & 4.6 & 7.0 & -- &  & -- &  & -- & -- & -- \\
    & DynaMoN~\cite{schischka2024dynamon} & \checkmark & 2.1 & 1.7 & \cellcolor{gray}\textbf{1.3} & 2.7 & 2.1 & 2.0 & 2.2 & 17.2 & 1.5 & \cellcolor{gray}\textbf{1.9} & 0.6\\
    & D$^2$GSLAM (Ours) & \checkmark & \cellcolor{gray}\textbf{1.5} & \cellcolor{gray}\textbf{1.6} & \cellcolor{gray}\textbf{1.3} & \cellcolor{gray}\textbf{2.6} & \cellcolor{gray}\textbf{1.8} & \cellcolor{gray}\textbf{1.9} & \cellcolor{gray}\textbf{2.1} & \cellcolor{gray}\textbf{7.1} & \cellcolor{gray}\textbf{1.4} & \cellcolor{gray}\textbf{1.9} & \cellcolor{gray}\textbf{0.5}\\
    \toprule
     \end{tabular}}
  \label{tab:bonn_low}
    \vspace{0.5em} 
    \tiny
    \raggedright 
    \textbf{*} denotes the reproduced results by running official code.  Results of other methods are directly taken from their papers. \textbf{--} denotes that no results were reported for these sequences. Best results are highlighted as \colorbox{gray}{\textbf{first}}.
\end{table*}

\begin{table*}[h]
  \centering
 \caption{Comparison of our method with SLAM baselines in tracking performance. We report ATE RMSE [cm]~$\downarrow$ on \textbf{8 dynamic scenes} of TUM dataset. (DM) methods that perform dynamic modeling.   
 }
  \resizebox{1.0\linewidth}{!}{
  \begin{tabular}{l|l |c |c|c |c|c |c|c |c|c}
    \toprule
    & \multirow{2}{*}{Methods} & \multirow{2}{*}{DM} & \multicolumn{4}{c|}{High-Dynamic} & \multicolumn{4}{c}{Low-Dynamic}\\
    &  &  & f3/walk\_st & f3/walk\_xyz & f3/walk\_half & f3/walk\_rpy & f3/sit\_st & f3/sit\_xyz & f3/sit\_half & f3/sit\_rpy \\
    \midrule
    \multirow{4}{*}{\shortstack{Static\\SLAM}}
    & SplaTAM*~\cite{keetha2024splatam} & & 115.2 & 218.3 & 68.4 & 100.4 & 7.1 & 1.7 & 28.1 & 23.4\\
    & MonoGS*~\cite{matsuki2024gaussian} & & 10.5 & 29.4 & 54.1 & 33.9 & 1.1 & 2.7 & 7.4 & 13.8\\
    & GS-ICP SLAM*~\cite{ha2025rgbd} & & 95.1 & 80.6 & 76.7 & 149.5 & 1.7 & 3.6 & 10.3 & 9.6\\
    & DROID-SLAM*~\cite{teed2021droid} & &  1.4 & 12.2 & 15.0 & 6.1 & 0.6 & 8.0 & 10.6 & 2.4\\
     \midrule
    \multirow{9}{*}{\shortstack{Dynamic\\SLAM}}
    & ReFusion~\cite{palazzolo2019refusion} & & 1.7 & 9.9 & 10.4 & -- & 0.9 & 4.0 & 11.0 & --\\
    & StaticFusion~\cite{scona2018staticfusion} & & 1.4 & 12.7 & 6.3 & -- & 1.3 & 4.0 & 4.0 & --\\
    & DynaSLAM~\cite{bescos2018dynaslam}] & & \cellcolor{gray}\textbf{0.6} & 1.5 & 2.5 & 3.5 & 0.7 & 1.5 & 1.7 & --\\
    & DDN-SLAM~\cite{li2024ddn}& & 1.0 & \cellcolor{gray}\textbf{1.4} & 2.3 & 3.9 & -- & 1.0 & 1.7 & --\\
    & RoDyn-SLAM*~\cite{jiang2024rodyn} & & 1.7 & 8.3 & 5.6 & 7.8 & 1.0 & 5.1 & 4.4 & 4.5\\
    & LMF-SLAM~\cite{li2024learn} & & 2.5 & 7.6 & 7.9 & -- & 0.7 & 1.8 & 3.9 & --\\
    & DG-SLAM~\cite{xu2025dg} & & \cellcolor{gray}\textbf{0.6} & 1.6 & -- & 4.3 & -- & 1.0 & -- & --\\
    & PG-SLAM~\cite{li2025pg} & \checkmark & 1.4 & 6.8 & 11.7 & -- & 0.7 & 1.5 & 4.0 & 5.4\\
    & DynaMoN~\cite{schischka2024dynamon}& \checkmark &  0.7 & \cellcolor{gray}\textbf{1.4} & 1.9 & 3.1 & 0.5 & 1.0 & 2.3 & 2.4\\
    & D$^2$GSLAM (Ours) & \checkmark & \cellcolor{gray}\textbf{0.6} & \cellcolor{gray}\textbf{1.4} & \cellcolor{gray}\textbf{1.8} & \cellcolor{gray}\textbf{3.0} & \cellcolor{gray}\textbf{0.4} & \cellcolor{gray}\textbf{0.9} & \cellcolor{gray}\textbf{1.4} & \cellcolor{gray}\textbf{2.3}\\
    \toprule
     \end{tabular}}
  \label{tab:tum}
   \vspace{0.5em} 
    \tiny
    \raggedright 
    \textbf{*} denotes the reproduced results by running official code.  Results of other methods are directly taken from their papers. \textbf{--} denotes that no results were reported for these sequences. Best results are highlighted as \colorbox{gray}{\textbf{first}}.
\end{table*}

\begin{table*}[h]
  \centering
  \caption{Comparison of our method with SLAM baselines in tracking performance. We report ATE RMSE [cm]~$\downarrow$ on \textbf{10 static scenes} of TUM and ScanNet dataset.}
  \resizebox{1\linewidth}{!}{
  \begin{tabular}{c| l |cc cc c|cc cc c}
    \toprule
    & \multirow{2}{*}{Methods} & \multicolumn{5}{c|}{TUM} & \multicolumn{5}{c}{ScanNet}\\
    & & f1/xyz & f1/rpy & f1/desk & f2/xyz & f3/office & scene0000 & scene0059 & scene0169 & scene0181 & scene0207\\
    \midrule
    \multirow{3}{*}{\shortstack{Static\\SLAM}}
    & SplaTAM*~\cite{keetha2024splatam} & 2.6 & 5.1 & 3.4 & 1.2 & 5.2 & 12.8 & 10.1 & 12.1 & 11.1 & 7.5\\
    & MonoGS*~\cite{matsuki2024gaussian} & \cellcolor{gray}\textbf{1.0} & 3.9 & \cellcolor{gray}\textbf{1.5} & 1.4 & 1.5 & 23.0 & 8.1 & 13.7 & 13.4 & 19.4\\
    & GS-ICP SLAM*~\cite{ha2025rgbd} & 1.5 & 3.2 & 2.7 & 1.8 & 2.7 & 40.4 & 38.6 & 42.3 & 21.9 & 14.1\\
     \midrule
    \multirow{2}{*}{\shortstack{Dynamic\\SLAM}}
    & DG-SLAM~\cite{xu2025dg} & -- & -- & -- & -- & -- & 7.9 & 11.5 & 8.3 & \cellcolor{gray}\textbf{7.3} & 8.2\\
    & D$^2$GSLAM (Ours) & \cellcolor{gray}\textbf{1.0} &  \cellcolor{gray}\textbf{2.0} &  \cellcolor{gray}\textbf{1.5} &  \cellcolor{gray}\textbf{0.2} &  \cellcolor{gray}\textbf{1.4} &  \cellcolor{gray}\textbf{5.6} &  \cellcolor{gray}\textbf{7.7} &  \cellcolor{gray}\textbf{8.2} &  \cellcolor{gray}\textbf{7.3} &  \cellcolor{gray}\textbf{6.5} \\
    \toprule
     \end{tabular}}
    \label{tab:static}
    \vspace{0.5em} 
    \tiny
    \raggedright 
    \textbf{*} denotes the reproduced results by running official code.  Results of other methods are directly taken from their papers. \textbf{--} denotes that no results were reported for these sequences.  Best results are highlighted as \colorbox{gray}{\textbf{first}}. 
\end{table*}

\section{Experiments}
\subsection{Experimental Setup}
\textbf{(1) Datasets:}
We evaluate the performance of our D$^2$GSLAM on 30 dynamic scenes of Bonn RGB-D dataset~\cite{palazzolo2019refusion} and TUM RGB-D dynamic dataset~\cite{sturm2012benchmark}. Moreover, we also assess the accuracy of our method on 10 static scenes of TUM RGB-D static dataset~\cite{sturm2012benchmark} and ScanNet dataset~\cite{dai2017scannet}. This selection enables comprehensive evaluation of our method across varied scenarios, validating its robustness in real-world indoor environments.

\noindent\textbf{(2) Metrics:}
For 3D reconstruction evaluation, we use \textit{F-score(\%)}, \textit{Depth L1(cm)}, \textit{Accuracy(cm)}, \textit{Completion(cm)}, and \textit{Completion ratio(\%)} metrics.
Moreover, we employ \textit{PSNR (dB)}, \textit{SSIM}, and \textit{LPIPS} metrics to evaluate the rendering quality of dynamic reconstruction.
Also, we report ATE~\cite{sturm2012benchmark} RMSE metric for tracking accuracy evaluation. 
\begin{itemize}
    \item F-score(\%): \(F=2\frac{PR}{P + R}\), where precision (\(P\)) measures the percentage of points on the predicted mesh that are within 1 cm distance to the ground truth mesh, and recall (\(R\)) vice versa.
    \item Depth L1(cm): the average L1 loss between ground truth depth and rendered depth. 
    \item Accuracy(cm): the average distance from sampled points on the reconstructed mesh to their nearest ground truth points.
    \item Completion(cm): the average distance from sampled points on the ground truth mesh to their nearest points on the reconstructed mesh.
    \item Completion ratio(\%): the percentage of points in the reconstructed mesh with completion under 5 cm.
\end{itemize}

\noindent\textbf{(3) Baselines:}
We compare our method with traditional dynamic SLAM~\cite{palazzolo2019refusion, scona2018staticfusion, bescos2021dynaslam}, NeRF-based dynamic SLAM~\cite{li2024ddn, li2024learn, jiang2024rodyn, schischka2024dynamon}, 3DGS-based dynamic SLAM~\cite{xu2025dg, li2025pg}, and static SLAM systems~\cite{keetha2024splatam, matsuki2024gaussian, ha2025rgbd, teed2021droid} which are not designed for dynamic environments. Note that we report DynaSLAM (N+G) version of \cite{bescos2021dynaslam} and DynaMoN (MS\&SS) version of \cite{schischka2024dynamon}.
However, most dynamic SLAM methods focus on removing dynamic objects to reconstruct static scenes and fail to perform dynamic modeling, which makes them infeasible for dynamic reconstruction evaluation.
DynaMoN~\cite{schischka2024dynamon} achieves dynamic modeling in an offline manner, requiring approximately 10 hours of training for one scene, while our method takes only tens of minutes for dynamic reconstruction. 
Therefore, for a fair comparison, we only compare our method with online SLAM systems on reconstruction performance, including existing 3DGS-based SLAM~\cite{keetha2024splatam, matsuki2024gaussian, ha2025rgbd}. 

\noindent\textbf{(4) Implementation Details:}
We run D$^2$GSLAM on a desktop with a single NVIDIA RTX 4090 GPU. For static parts of the scene, we initialize 3D Gaussians using downsampled pixels for compact representation. 
We apply a downsampling rate of 32 during initialization. For the mapping process, we downsample the static background at a rate of 128.
Meanwhile, to maintain sufficient details in dynamic reconstruction, we initialize dynamic 4D Gaussians for each pixel in dynamic regions, using a downsampling rate of 1. 
During optimization, we perform 150 iterations for each keyframe to optimize the static background, while the dynamic components are refined through a per-frame retrospective optimization process with 100 iterations.
The weighting coefficients of each loss are $\lambda_{c}=0.9$, $\lambda_{d}=0.1$, $\lambda_{sp}=0.05$, $\lambda_{mc}=0.1$, $\lambda_c^t=0.9$, $\lambda_d^t=0.1$.
For dynamic separation process, $\tau_d=0.5$, $\sigma_d=400$.

We use Adam optimizer to optimize Gaussian parameters. For optimizing static 3D Gaussians, we use a learning rate of 1.6$e^{-6}$ for 3D center position optimization, 0.001 for scales optimization, 0.001 for rotation optimization, 0.05 for opacity optimization, and 0.0025 for color optimization. In terms of dynamic 4D Gaussians, we use a learning rate of 8$e^{-4}$ for 3D center position optimization, 1.6$e^{-4}$ for time optimization, 0.005 for both 3D scales and time scales optimization, 0.05 for opacity optimization, and 0.0025 for color optimization.

\begin{table*}[h]
  \centering
     \caption{Comparison of our method with dynamic SLAM baselines in 3D mapping performance. We report \textit{Accuracy}[cm] $\downarrow$,  \textit{Completion}[cm] $\downarrow$, \textit{Completion Ratio}[\%] $\uparrow$, \textit{F-score}[\%] $\uparrow$, \textit{Depth L1}[cm] $\downarrow$. 
     }
  \resizebox{1\linewidth}{!}{
  \begin{tabular}{l| ccccc | ccccc | ccccc | ccccc}
    \toprule
    \multirow{2}{*}{Methods} & \multicolumn{5}{c|}{balloon} & \multicolumn{5}{c|}{balloon2} & \multicolumn{5}{c|}{person\_tracking} & \multicolumn{5}{c}{person\_tracking2}\\
    & Acc. & Com. & Com.R & F-s & D\_L1 & 
    Acc. & Com. & Com.R & F-s & D\_L1 &
    Acc. & Com. & Com.R & F-s & D\_L1 &
    Acc. & Com. & Com.R & F-s & D\_L1 \\
    \midrule
    ReFusion~\cite{palazzolo2019refusion} & 8.2 & 12.6 & 31.6 & -- & -- & 
    7.9 & 11.7 & 32.2  & -- & -- &
    46.9 & 104.0 & 13.9  & -- & -- &
    78.5 & 166.6 & 10.6 & -- & --\\
    DG-SLAM~\cite{xu2025dg} & 7.0 & 9.8 & 49.5 & -- & -- &
    5.8 & 8.1 & 52.4 & -- & -- &
    9.1 & 18.0 & 34.6 & -- & -- &
    11.8 & 20.1 & 32.8 & -- & -- \\
    RoDyn-SLAM*~\cite{jiang2024rodyn} & 10.6 & 7.2 & 47.6 & 41.7 & 66.8&
    13.4 & 7.9 & 40.9 & 38.1 & 68.4 &
    10.2 & 27.7 & 34.1 & 30.1 & 78.8 &
    13.8 & 19.0 & 32.6 & 24.3 & 80.6\\
    D$^2$GSLAM (Ours) & \cellcolor{gray}\textbf{6.8} & \cellcolor{gray}\textbf{6.2} & \cellcolor{gray}\textbf{57.3} & \cellcolor{gray}\textbf{49.8} & \cellcolor{gray}\textbf{2.8} &
     \cellcolor{gray}\textbf{4.3} & \cellcolor{gray}\textbf{6.0} & \cellcolor{gray}\textbf{61.2} & \cellcolor{gray}\textbf{57.8} & \cellcolor{gray}\textbf{2.9} & 
     \cellcolor{gray}\textbf{8.5} & \cellcolor{gray}\textbf{13.4} & \cellcolor{gray}\textbf{41.7} & \cellcolor{gray}\textbf{35.2} & \cellcolor{gray}\textbf{2.5} &
     \cellcolor{gray}\textbf{10.6} & \cellcolor{gray}\textbf{18.1} & \cellcolor{gray}\textbf{37.4} & \cellcolor{gray}\textbf{31.5} & \cellcolor{gray}\textbf{2.6} \\
    \toprule
     \end{tabular}}
  \label{tab:bonn_recon}
   \vspace{0.5em} 
   \tiny
   \raggedright 
   \textbf{*} denotes the reproduced results by running official code.  Results of other methods are directly taken from their papers. \textbf{--} denotes that no results were reported for these sequences. Best results are highlighted as \colorbox{gray}{\textbf{first}}. 
\end{table*}

\begin{table*}[h]
  \centering
    \caption{Quantitative comparison of rendering performance on Bonn and TUM dataset. 
    }   
  \resizebox{\linewidth}{!}{%
  \begin{tabular}{l|c| ccccc|ccc|c}
    \toprule
    \multirow{2}{*}{Methods} & \multirow{2}{*}{Metrics} & \multicolumn{5}{c|}{Bonn} & \multicolumn{3}{c|}{TUM} & \multirow{2}{*}{Avg.} \\
    & & balloon & balloon2 & crowd & syn & person\_track & fr3/walk\_static& fr3/walk\_xyz & fr3/sit\_xyz\\
    \midrule
    \multirow{3}{*}{SplaTAM*~\cite{keetha2024splatam}} & PSNR[dB] $\uparrow$ & 19.83 & 17.73 & 17.35 & 19.34 & 16.28 & 19.09 & 16.82 & 22.13 & 18.57\\
     & SSIM $\uparrow$ & 0.783 & 0.699 & 0.656 & 0.771 & 0.622 & 0.717 & 0.640 & 0.880 & 0.721\\
     & LPIPS $\downarrow$ & 0.218 & 0.293 & 0.297 & 0.237 & 0.339 & 0.261 & 0.348 & 0.158 & 0.269\\
     \midrule
     \multirow{3}{*}{MonoGS*~\cite{matsuki2024gaussian}} & PSNR[dB] $\uparrow$ & 18.81 & 17.91 & 15.22 & 19.04 & 14.57 & 14.24 & 13.67 & 20.68 & 16.77 \\
     & SSIM $\uparrow$ & 0.721 & 0.686 & 0.601 & 0.671 & 0.651 & 0.521 & 0.422& 0.744 & 0.627\\
     & LPIPS $\downarrow$ & 0.444 & 0.454 & 0.582 & 0.452 & 0.584 & 0.414 & 0.544& 0.229 & 0.463\\
     \midrule
     \multirow{3}{*}{GS-ICP SLAM*~\cite{ha2025rgbd}} & PSNR[dB] $\uparrow$ & 16.30 & 16.00 & 16.85 & 17.65 & 16.99 & 18.21 & 17.31 & 21.41 & 17.59\\
     & SSIM $\uparrow$ & 0.741 & 0.712 & 0.725 & 0.716 & 0.745 & 0.721 & 0.706 & 0.780 & 0.731\\
     & LPIPS $\downarrow$ & 0.299 & 0.338 & 0.323 & 0.297 & 0.316 & 0.320 & 0.350 & 0.192 & 0.307\\
     \midrule
     \multirow{3}{*}{D$^2$GSLAM (Ours)} & PSNR[dB] $\uparrow$ &\cellcolor{gray}\textbf{26.16} & \cellcolor{gray}\textbf{24.58} & \cellcolor{gray}\textbf{24.13} & \cellcolor{gray}\textbf{25.24} & \cellcolor{gray}\textbf{24.88} & \cellcolor{gray}\textbf{22.42} & \cellcolor{gray}\textbf{22.19} & \cellcolor{gray}\textbf{23.14} & \cellcolor{gray}\textbf{24.09}\\
     & SSIM $\uparrow$ & \cellcolor{gray}\textbf{0.881} & \cellcolor{gray}\textbf{0.832} & \cellcolor{gray}\textbf{0.814} & \cellcolor{gray}\textbf{0.876} & \cellcolor{gray}\textbf{0.855} & \cellcolor{gray}\textbf{0.811} & \cellcolor{gray}\textbf{0.806} & \cellcolor{gray}\textbf{0.885} & \cellcolor{gray}\textbf{0.845}\\
     & LPIPS$\downarrow$ & \cellcolor{gray}\textbf{0.133} & \cellcolor{gray}\textbf{0.185} & \cellcolor{gray}\textbf{0.186} & \cellcolor{gray}\textbf{0.139} & \cellcolor{gray}\textbf{0.161} & \cellcolor{gray}\textbf{0.194} & \cellcolor{gray}\textbf{0.205} & \cellcolor{gray}\textbf{0.150} & \cellcolor{gray}\textbf{0.169}\\
     \toprule
      \end{tabular}%
    }       
    \label{tab:rendering}
    \vspace{0.5em} 
    \tiny
    \raggedright 
    \textbf{*} denotes the reproduced results by running official code. Best results are highlighted as \colorbox{gray}{\textbf{first}}. 
\end{table*}

\subsection{Experimental Results}
\subsubsection{Tracking}
We conduct experiments on high-dynamic, low-dynamic, and static scenes to assess the tracking performance of our D$^2$GSLAM.
High-dynamic scenarios refer to environments that contain fast-moving objects or multiple moving objects, such as flying balloons. Low-dynamic scenarios refer to scenes with relatively slow movements. Since tracking module of DynaMoN~\cite{schischka2024dynamon} operates online, we also compare our tracking performance with this method. 

As shown in Tab.~\ref{tab:bonn_high}, Tab.~\ref{tab:bonn_low}, and Tab.~\ref{tab:tum}, our method outperforms other SLAM methods across all 30 dynamic scenes, achieving up to 42\% increase in tracking accuracy.
Such improvement is attributed to our proposed geometric-prompt dynamic separation module, which leverages geometric consistency of 3DGS representation to effectively detect and mask dynamic objects. Consequently, our D$^2$GSLAM remains robust performance despite geometric inconsistencies caused by dynamic objects, leading to superior tracking precision.
Moreover, when moving from dynamic to static environments, our method maintains robust performance across all scenes on TUM and ScanNet datasets in Tab.~\ref{tab:static}.

\subsubsection{3D Reconstruction}
Tab.~\ref{tab:bonn_recon} compares the 3D reconstruction accuracy of our method with other dynamic SLAM baselines. 
While other dynamic SLAM methods typically remove dynamic objects and only reconstruct static scenes, our approach simultaneously reconstructs both dynamic objects and static backgrounds. For a fair comparison, we evaluate the reconstruction quality of our method only on the static background. 
D$^2$GSLAM achieves up to 96\% improvement in \textit{Depth L1} metric and average of 36\% improvement in reconstruction metrics.
This improvement is attributed to our dynamic-static composite mapping, which achieves precise dense reconstruction of the scene.

\begin{figure*}
    \centering
    \includegraphics[width=\linewidth]{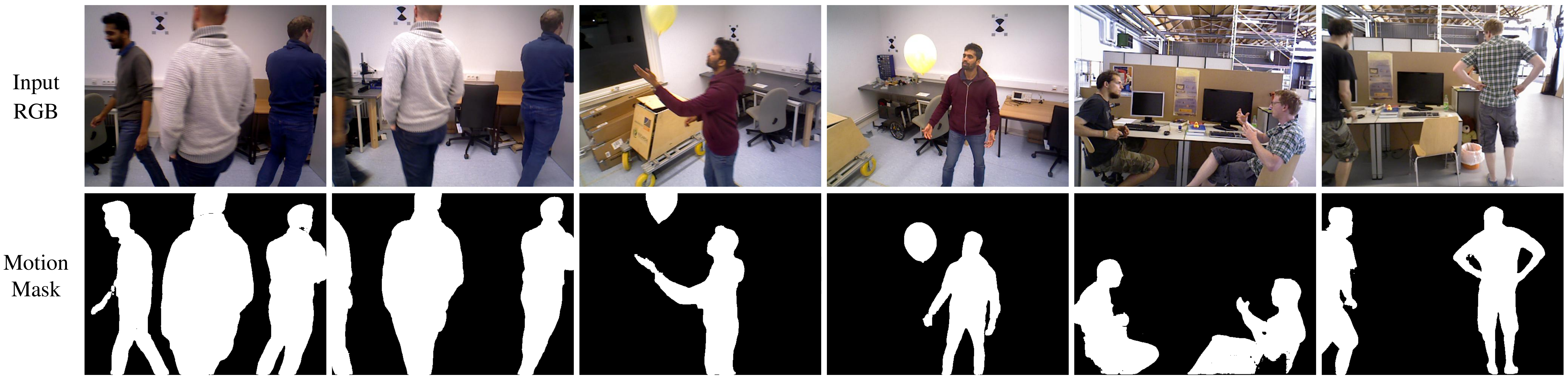}
    \caption{Visualization of our generated motion masks. Our method achieves accurate dynamic separation results in various dynamic scenes.}
    \label{fig:vis_mask}
\end{figure*}

\begin{figure*}
    \centering
    \includegraphics[width=\linewidth]{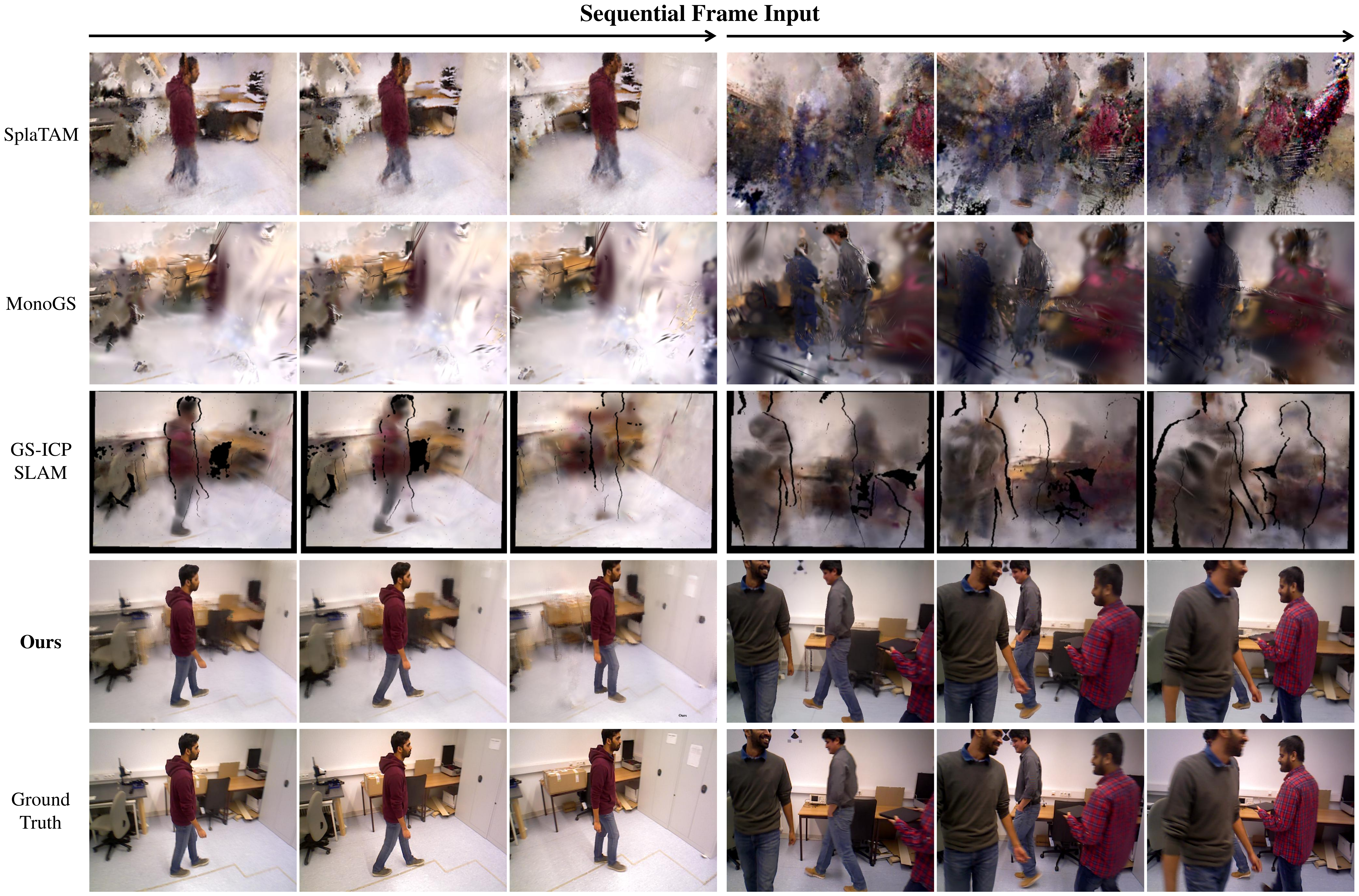}
    \caption{\textbf{Qualitative comparison of scene reconstruction performance on Bonn dataset}. People are walking in the scenes. Our method achieves high-quality dynamic reconstruction results in challenging dynamic scenarios, including complex scene with multiple people moving. 
    Other SLAM methods fail to perform mapping in such dynamic scenes.}
    \label{fig:dynamic_bonn}
\end{figure*}

\begin{figure*}
    \centering
    \includegraphics[width=\linewidth]{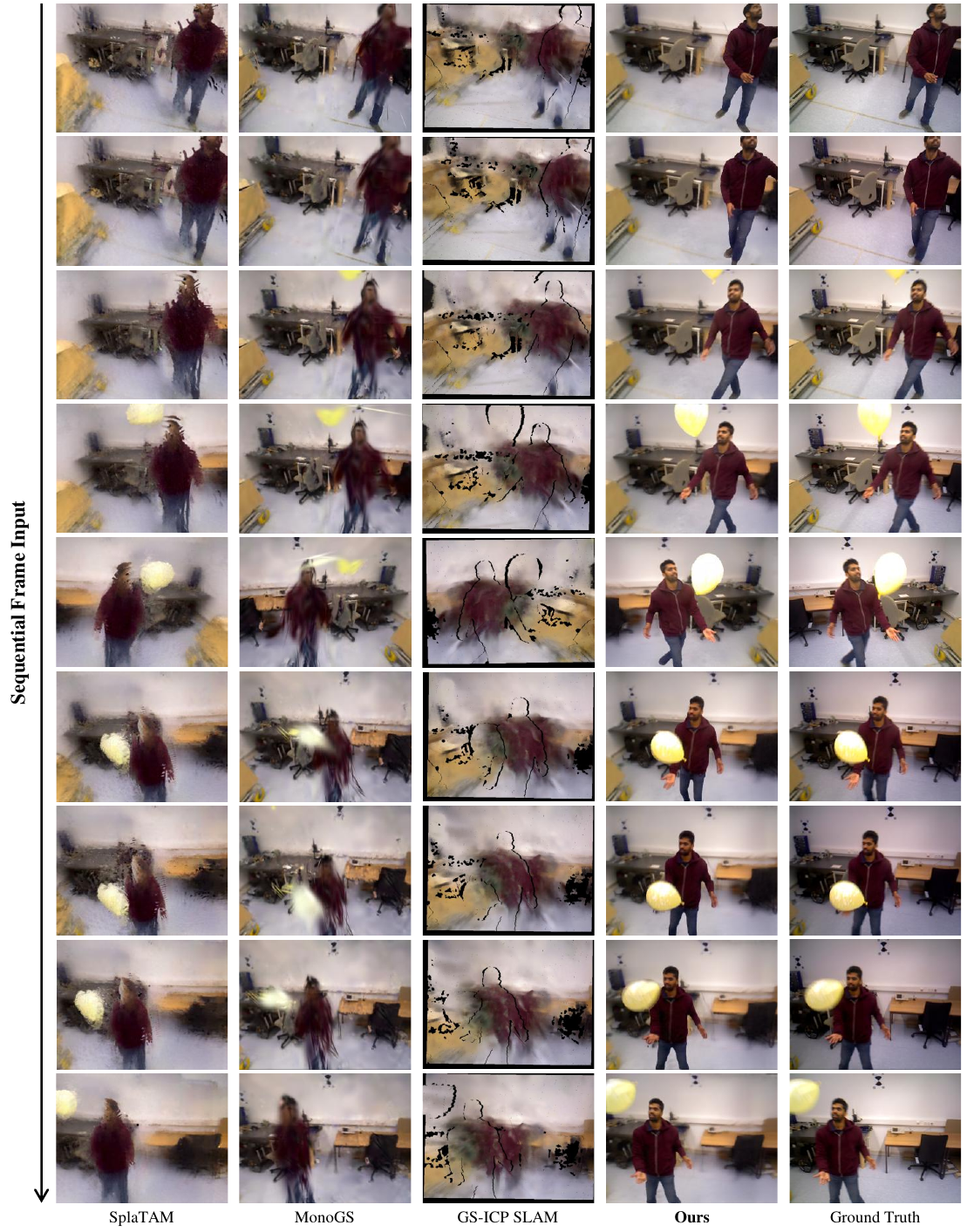}
    \caption{\textbf{Qualitative comparison of scene reconstruction performance}. The person is hitting a balloon with rapid up-and-down motions in the scene. Our method achieves accurate dynamic reconstruction compared with other SLAM methods.}
    \label{fig:balloon2_vis}
\end{figure*}

\begin{table}
    \centering
     \caption{Runtime comparison on Bonn dataset. (MS) motion separation.}
        \resizebox{\columnwidth}{!}{
        \begin{tabular}{l|ccc|c}
        \toprule
        \multirow{2}{*}{Methods} & MS & Track & Map & Total Time\\
        & (ms) & (ms) & (ms) & (ms)\\
        \midrule
        % DynaSLAM~\cite{bescos2021dynaslam} & 528 & \textbf{2} & --- & 530\\
        LMF-SLAM~\cite{li2024learn} & 628 & 149 & 161 & 938\\
        Rodyn-SLAM~\cite{jiang2024rodyn} & 279 & 876 & 1084 & 2239\\
        DG-SLAM~\cite{xu2025dg} & 163 & 89 & 549 & 801\\
        D$^2$GSLAM (Ours) & \cellcolor{gray}\textbf{88} & \cellcolor{gray}\textbf{68} & \cellcolor{gray}\textbf{202} & \cellcolor{gray}\textbf{358}\\
        \bottomrule
         \end{tabular}}
        \label{tab:runtime_compare}
\end{table}

\begin{table}
    \centering
    \caption{Detailed runtime of our motion separation process on Bonn dataset. Our motion separation consists of dynamic detector and motion mask generation.
    ($R_g$) Geometry Consistency; ($\mathbb{C}_g$): Connected Component Extraction; ($R_p$) Prediction Residual.
    }
    \resizebox{\columnwidth}{!}{
    \begin{tabular}{cc|ccc|c}
    \toprule
    \multicolumn{2}{c|}{Dynamic Detector} & \multicolumn{3}{c|}{Motion Mask Generation} & Total\\
    \cline{1-5}
    $R_g$  & $\mathbb{C}_g$  & $R_p$ & YOLO-world & MobileSAM & Time\\
    (ms) & (ms) & (ms) & \cite{cheng2024yolo} (ms) & \cite{zhang2023faster} (ms) & (ms)\\
    \midrule
    34.7 & 1.1 & 5.9 & 16.0 & 30.5 & 88.2\\
    \bottomrule
    \end{tabular}
    }
    \label{tab:runtime_detail}
\end{table}

\subsubsection{Runtime Analysis}
As shown in Tab.~\ref{tab:runtime_compare}, we compare per-frame runtime with other dynamic SLAM methods. Since other SLAM methods only reconstruct static backgrounds and directly remove dynamic objects, we solely report the runtime of motion separation, tracking, and static mapping for a fair comparison. 
Note that our reported motion separation time includes the combined runtime of both dynamic detector and motion mask generation module.
Our SLAM system runs 2 times faster than state-of-the-art dynamic SLAM methods, demonstrating superior computational efficiency while maintaining robust performance. In addition, for dynamic modeling runtime, each frame requires 870\textit{ms} of optimization for accurate modeling. Since we perform tracking only on static parts, the time consumption introduced by dynamic modeling does not affect the real-time performance of tracking and static reconstruction. Instead, we consider the dynamic modeling as back-end optimization for more accurate reconstruction of dynamic scenes. 

As our motion separation consists of several parts, we further provide a detailed runtime analysis of each part in Tab.~\ref{tab:runtime_detail}. Our D$^2$GSLAM only requires 88.2$ms$ to achieve an accurate motion mask. In contrast, other dynamic SLAM methods, such as LMF-SLAM~\cite{li2024learn} requires 628$ms$, Rodyn-SLAM~\cite{jiang2024rodyn} requires 279$ms$, DG-SLAM~\cite{xu2025dg} requires 163$ms$ to obtain a motion mask. Our motion segmentation is 2 times faster than other dynamic SLAM methods.
This real-time performance is attributed to our geometric-prompt separation strategy that fully leverages scene geometry consistency to obtain dynamic regions, which outperforms other motion segmentation methods in terms of both efficiency and accuracy.

\subsubsection{Dynamic Separation}
Fig.~\ref{fig:vis_mask} shows that our generated motion masks achieve accurate dynamic separation results in various dynamic scenes. Specifically, our method  can distinguish the dynamic regions of people leaving the viewing frame, fast-moving balloons, and detailed parts of moving objects such as human hands. This capability is attributed to our proposed geometric-prompt dynamic separation method, which leverages the geometric consistency of scene representation for precise motion segmentation.

\subsubsection{Dynamic Reconstruction}
As shown in Tab.~\ref{tab:rendering}, our method significantly surpasses baselines across all dynamic scenes, showing accurate dynamic reconstruction capability. 
Meanwhile, we compare our reconstruction results with other SLAM methods in Fig.~\ref{fig:dynamic_bonn} and Fig.~\ref{fig:balloon2_vis}. Our method is capable of accurately modeling the motion of objects in the scene, such as people moving and a balloon rising, as well as achieving precise static background reconstruction. This improvement is due to our dynamic-static composite representation and retrospective frame optimization, which enables the complete reconstruction of both dynamic objects and static background through continuously revisiting and optimizing previously observed frames.

\begin{figure}
    \centering
    \includegraphics[width=\linewidth]{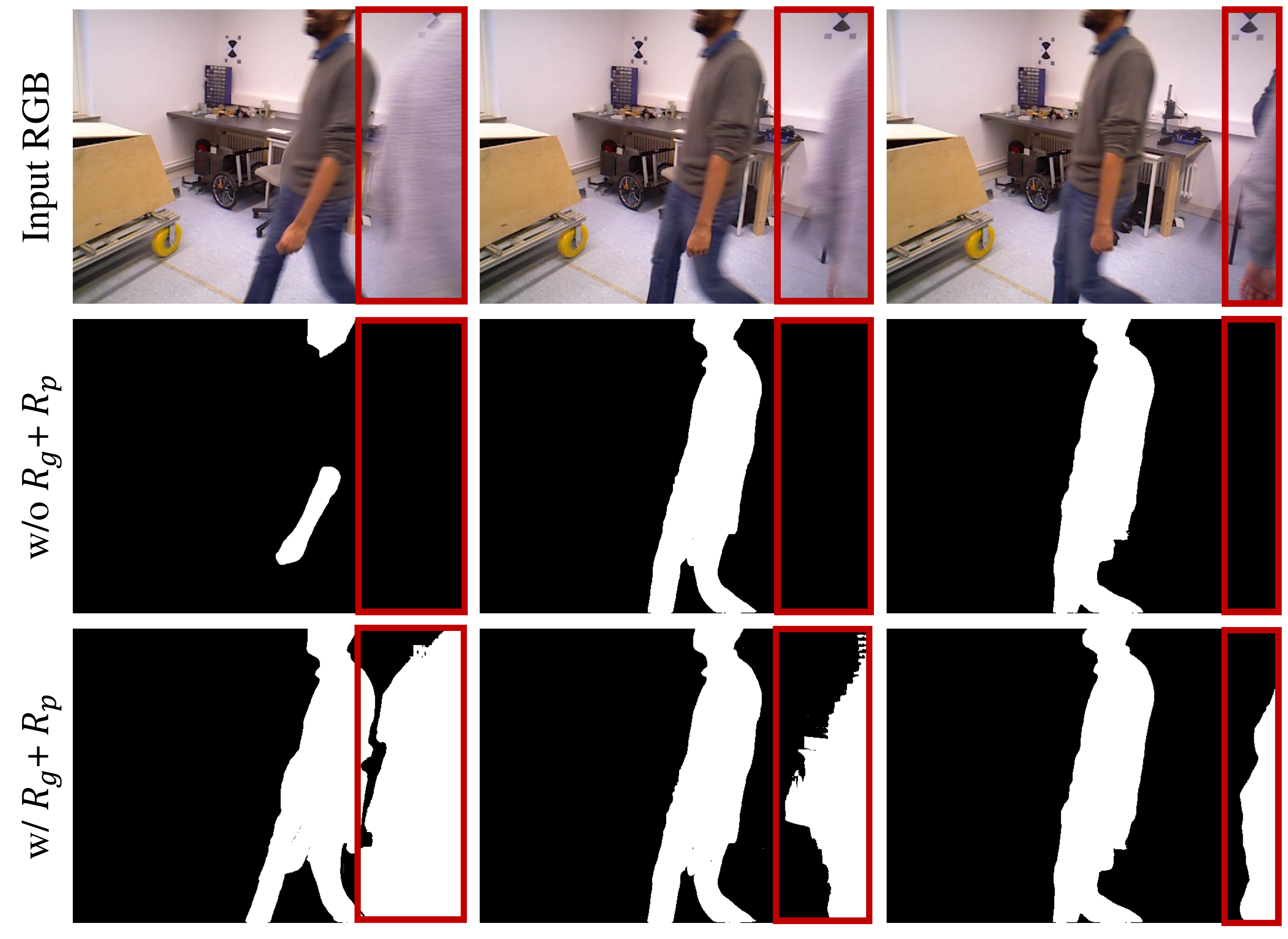}
    \caption{Ablation study of geometric-prompt dynamic separation module on Bonn dataset. (w/o $R_g + R_p$) without adding geometry consistency residual $R_g$ and prediction residual $R_p$.}
    \label{fig:ablation_mask}
\end{figure}

\subsection{Ablation Study}
We perform ablation study of our D$^2$GSLAM, which consists of geometric-prompt dynamic separation followed by dynamic-static composite mapping and tracking. We first conduct ablation of our designed dynamic separation module. Then, we evaluate the effectiveness of retrospective frame optimization and motion consistency loss in composite mapping.

\noindent\textbf{Geometric-Prompt Dynamic Separation}\hspace{5pt}
We introduce geometry consistency residual $R_g$ and prediction residual $R_p$, which are combined with YOLO-world~\cite{cheng2024yolo} and MobileSAM~\cite{zhang2023faster} to obtain motion mask. As shown in Fig.~\ref{fig:ablation_mask}, when our designed $R_g$ and $R_p$ are not incorporated, the SLAM system fails to distinguish incomplete dynamic objects in the scene, such as people leaving the viewing frame. Additionally, when using only YOLO-world~\cite{cheng2024yolo} and MobileSAM~\cite{zhang2023faster} for dynamic separation, the system may fail to generate a complete motion mask of the people, even when they are in the center of the image. In contrast, by incorporating our proposed $R_g$ and $R_p$, we can effectively identify various dynamic objects and generate complete motion masks for these objects in the scene. The improved performance arises from $R_g$ and $R_p$ leveraging the geometry consistency of Gaussian representation and scene geometry, which enables precise distinction of dynamic objects through spatial inconsistency.

\begin{figure}
    \centering
    \includegraphics[width=\linewidth]{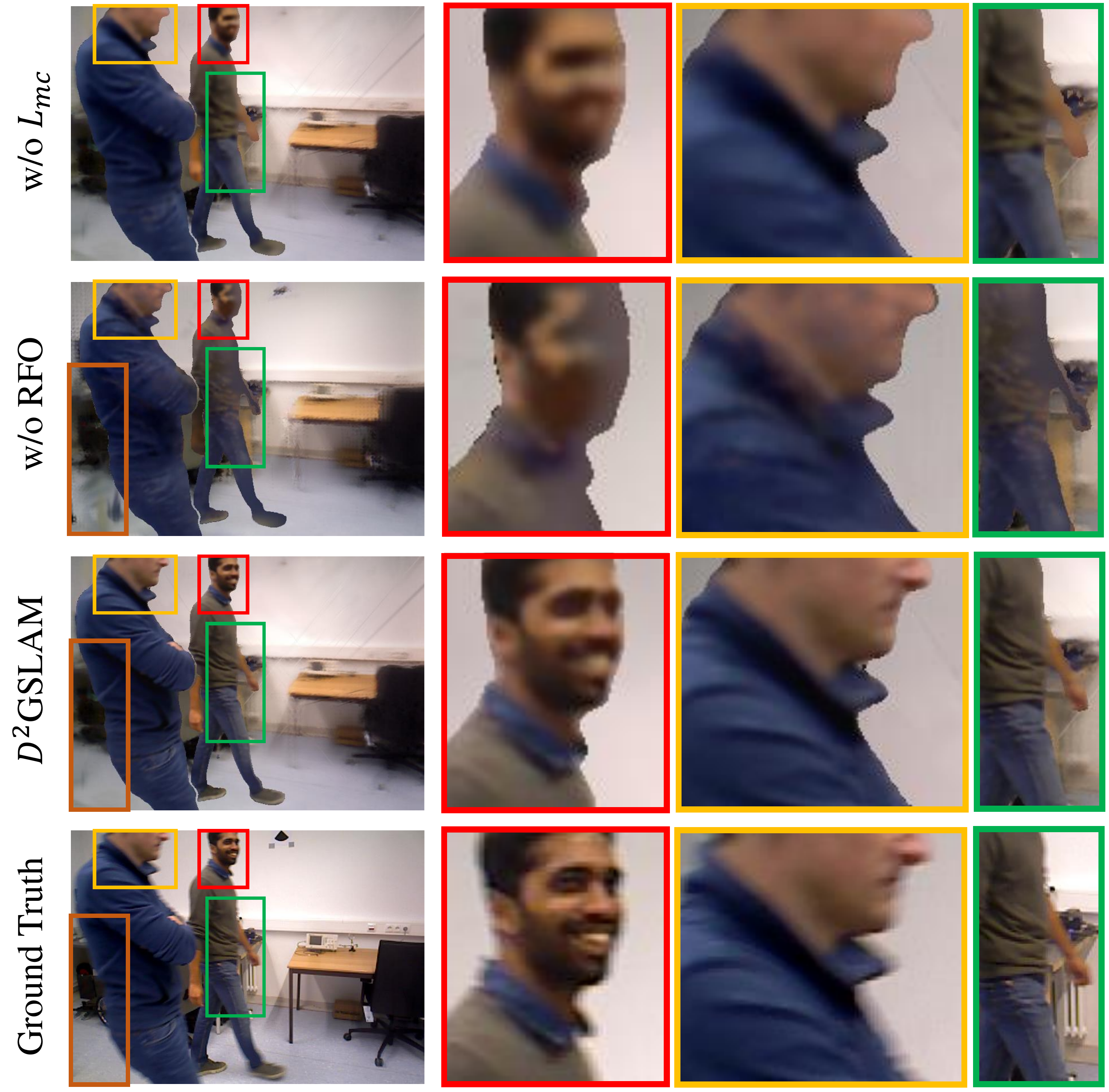}
    \caption{Ablation study of dynamic reconstruction on Bonn dataset.  (w/o RFO) without retrospective frame optimization; (w/o $\mathcal{L}_{mc}$) without motion consistency loss. It can be seen that the whole D$^2$GSLAM achieves the best dynamic reconstruction accuracy. Details are highlighted with colorful boxes, where \textcolor{myorange}{orange box} shows static background and other color boxes are details of dynamic objects.}
    \label{fig:ablation_dynamic}
\end{figure}

\noindent\textbf{Retrospective Frame Optimization}\hspace{5pt}
As shown in Fig.~\ref{fig:ablation_dynamic}, without Retrospective Frame Optimization (RFO), the SLAM system tends to forget previously modeled details, such as facial features, arms, and clothing details. In contrast, by incorporating RFO, our SLAM system achieves accurate dynamic modeling through continuously revisiting and optimizing previous observations. This strategy enables consistency of object motion modeling in both temporal and spatial domains.
In addition to enhancing dynamic reconstruction, RFO can also reduce catastrophic forgetting of static backgrounds. Fig.~\ref{fig:ablation_dynamic} shows that the previously reconstructed scene, such as the background behind the person in blue, becomes blurry during subsequent frame reconstruction without RFO. By applying RFO, our D$^2$GSLAM is capable of achieving complete scene reconstruction.

\noindent\textbf{Motion Consistency Loss}\hspace{5pt}
Fig.~\ref{fig:ablation_dynamic} shows that incorporating motion consistency loss $\mathcal{L}_{mc}$ enables more accurate reconstruction of dynamic object details during motion, such as facial features, arm movements, and clothing textures. 
This improvement is attributed to $\mathcal{L}_{mc}$ leveraging motion continuity to optimize dynamic scene representation, allowing better modeling of continuous object movements.

%% file: sec/5_conclusion.tex
\section{Conclusion}
\label{sec:conclusion}
We have proposed a novel dynamic SLAM system utilizing Gaussian representation, which achieves accurate dynamic reconstruction, motion modeling, and robust tracking in dynamic environments. We introduce geometric-prompt dynamic separation strategy to distinguish dynamic objects by leveraging geometry consistency of scene representation. 
Moreover, we propose a dynamic-static composite representation that incorporates static 3D Gaussians with dynamic 4D Gaussians for precisely reconstructing dynamic scenes.
Additionally, we employ a progressive pose refinement strategy that leverages both the multi-view consistency of static scene geometry and motion information from dynamic objects to achieve accurate camera tracking in dynamic scenes.
Furthermore, we introduce a motion consistency loss that utilizes temporal continuity of motions to achieve accurate dynamic modeling.
Our system has immediate applications in mixed reality and human-robot interaction scenarios, where accurate 3D and 4D mapping of dynamic environments is essential for effective operation.
Moreover, our research establishes a foundation for future advancements in motion prediction and unified scene understanding in dynamic scenes.

A main limitation of D$^2$GSLAM lies in its inability to achieve real-time dynamic modeling. While we consider the dynamic modeling as back-end optimization for more accurate reconstruction of dynamic scenes, real-time capability is crucial for SLAM applications, particularly in robotic scenarios where low-latency processing of dynamic changes is essential.
In future work, we aim to develop efficient real-time dynamic modeling techniques to enable practical deployment in robotic scenarios.